\documentclass[runningheads]{llncs}
\usepackage{graphicx}

\usepackage{tikz}
\usepackage{comment}
\usepackage{amsmath,amssymb} 
\usepackage{color, colortbl}
\usepackage{xspace}
\usepackage{arydshln}
\usepackage{svg}
\usepackage{hyperref}
\hypersetup{
  colorlinks,
  }

\usepackage[accsupp]{axessibility}  

\definecolor{lightgray}{gray}{.9}
\newif\ifcompilegraphics

\usepackage{enumitem}
\usepackage{multirow}

\usepackage{pifont}
\newcommand{\cmark}{\ding{51}}%

\compilegraphicstrue

\begin{document}
\pagestyle{headings}
\mainmatter

\title{On the Design of Privacy-Aware Cameras: \\a Study on Deep Neural Networks} 

\titlerunning{On the Design of Privacy-Aware Cameras}
%
\author{Marcela Carvalho \and
Oussama Ennaffi \and
Sylvain Chateau \and \\
Samy Ait Bachir}
\authorrunning{M. Carvalho et al.}
%
\institute{Upciti \\
}
\maketitle

\begin{abstract}
  In spite of the legal advances in personal data protection, the issue of private data being misused by unauthorized entities is still of utmost importance.
  To prevent this, Privacy by Design is often proposed as a solution for data protection.
  In this paper, the effect of camera distortions is studied using Deep Learning techniques commonly used to extract sensitive data.
  To do so, we simulate out-of-focus images corresponding to a realistic conventional camera with fixed focal length, aperture, and focus, as well as grayscale images coming from a monochrome camera.
  We then prove, through an experimental study, that we can build a privacy-aware camera that cannot extract personal information such as license plate numbers.
  At the same time, we ensure that useful non-sensitive data can still be extracted from distorted images.
  Code is available on~\href{https://github.com/upciti/privacy-by-design-semseg}{https://github.com/upciti/privacy-by-design-semseg}

\end{abstract}

\section{Introduction}

With the propagation of cameras in the public space, each year with higher definition, data privacy has become a general concern.
We are in fact surrounded by surveillance systems and high resolution cameras that generate billions of images per day~\cite{bogdanchikov2019face}.
These images can be processed using Computer Vision (CV) techniques in order to perform tasks such as object detection~\cite{he2017maskrcnn}, semantic segmentation~\cite{chen2017deeplabv3}, or object tracking~\cite{wang2019fast}.
However, and most importantly, these same CV techniques can be used to extract privacy sensitive information, such as faces~\cite{schroff2015facenet} or license plates~\cite{chang2004automatic} recognition.
The issue becomes even more problematic when these processes are performed without the notice or consent of consumers.

\begin{figure}[t]
  \centering
  \includegraphics[width=1.0\textwidth]{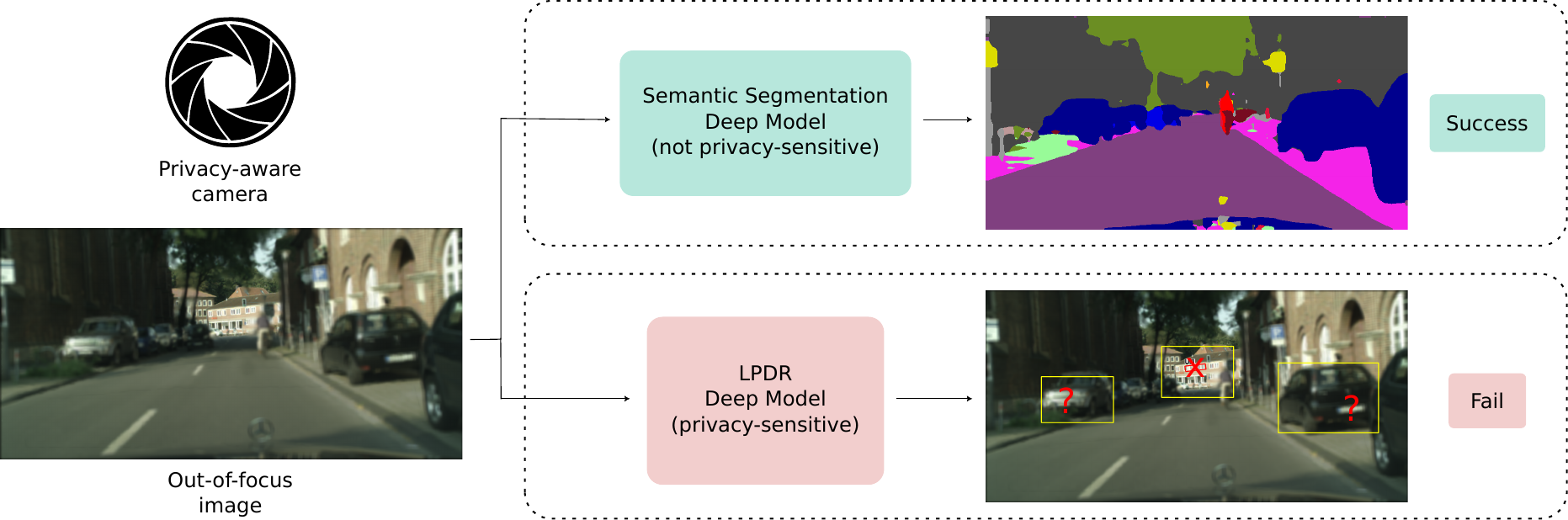}
  \caption{\label{fig:graphical-abstract} Overview of our method to ensure Privacy by Design. We propose the design of a hardware-level solution in which a non privacy-sensitive task achieves high quality results despite having a distorted image as input, while a privacy-sensitive task fails completely.}
\end{figure}

This situation raises many legal issues related to data protection.
As a result, a great deal of countries have adopted strict legislation to tackle this problematic~\cite{unctad2020data}.
For example, the European Union General Data Protection Regulation (GDPR) ensures confidentiality protection and transparency on the collection, storage and use of the data.
However, there exist techniques for anonymizing sensitive information in visual tasks. Traditional approaches consist in blurring, masking or changing pixels corresponding to personal identifiers~\cite{hukkelaas2019deepprivacy}, such as faces or license plate numbers.
Sadly, although this type of solution is GDPR compliant, it is prone to error and there are no guaranties that sensitive data cannot be compromised.
Additionally, usually owing to poor security, the cameras are not immune from being accessed directly by malicious people, thus putting in jeopardy the private personal data they record.

In view of these issues, Privacy by Design (PbD) seems to be a promising approach for the protection of private data~\cite{cavoukian2009privacy}.
Its principle is to guarantee personal information protection since the very conception of the devices.
In our particular case, when dealing with computer vision tasks, PbD can be achieved by limiting the visual quality with respect to the sensor capabilities. Indeed, by restricting the right portion of data acquired by the sensor, it is possible to highly limit the efficiency of privacy sensitive processing tasks while, in the same time, having low impact on the non sensitive ones.

In this work, we explore the feasibility of privacy-aware cameras that respect the PbD principle by studying their impacts on CV processing tasks which are usually applied to images taken in the public sphere.
To do so, we simulate the behavior of such a privacy-aware camera with two main visual distortions.
The first consists in generating defocus blur so that the images correspond to the outputs of a specially designed camera with fixed focal length, aperture and position of the in-focus plane; while the second distortion mimics a simple monochrome camera.
Grayscale images may improve anonymisation as it reduces contextual information related to specific color characteristics of personal objects. 
We then study the effects of such distortions on Deep Neural Networks (DNN) applied to several CV tasks.
These latter ones are classified into privacy-sensitive tasks, such as license plate recognition and non privacy-sensitive tasks, like object detection or semantic segmentation.
Our goal is thus to find out if there is a level of distortion that allows to process images automatically in an efficient way while preserving the privacy of people.

The rest of this paper is organised as follows.
Section~\ref{sec:related_work} presents background information related to the existing literature.
Section~\ref{sec:experiments} shows how the chosen image distortions influence the performance of the CNNs, depending on the sensitivity to details that we relate to privacy information.

\section{Related Work}~\label{sec:related_work}

Recent works in Deep Learning explore the impact of visual deformations on the robustness of DNNs~\cite{dodge2016understanding,hendrycks2018benchmarking,wang2021data,zhou2017classification,rusak2020simple}.
In~\cite{dodge2016understanding}, Dodge and Karam study the influence of 5 different quality distortions on neural networks for image classification and conclude they are specially sensitive to blur and noise.
Similarly, in~\cite{hendrycks2018benchmarking}, Hendrycks and Dietterich create a new dataset with 15 transformations from Imagenet~\cite{deng2009imagenet}, including Gaussian noise, defocus blur and fog. Then, they perform several tests on classification networks to explore their performance on out-of-distribution data and indicate which architectures are more robust to adversarial perturbations.
Other works include similar experiments with different colorizations~\cite{de2021impact} and blur~\cite{zhou2017classification}.
To overcome these limitations, some proposed approaches include these distortions in data augmentation to improve model efficiency~\cite{wang2021data,zhou2017classification,rusak2020simple}.
Their results and considerations improve our understanding on the deep models and also allow to overcome some of the observed constraints.
However, we may also use this intake to balance how much of information one can learn from the input images depending of the task to perform.

Previous works on data anonymisation can be divided in two branches: software-level and hardware-level anonymisation.
The first group of methods are more commonly adopted as they consist on a pre-processing phase and do not demand a specific camera~\cite{besmer2009tagged,vishwamitra2017blur,vishwamitra2017towards,ishii2020privacy}. 
They consist in automatically filtering sensitive information (\textit{e.g.}, facial features, license plates) by blurring or masking them, while preserving realistic background. More refined works in face recognition, like~\cite{hukkelaas2019deepprivacy}, propose to generate realistic faces to mask the original information while keeping data distribution. However, as mentioned in the last section, all of these methods are prone to error and if an attacker succeeds to access the code, the privacy of the method can be highly compromised~\cite{winkler2014security}.

Very few works propose a hardware-level solution to this concern.
This includes chip implementations with low-level processing arguably a type of software-level implementation~\cite{fernandez2014focal}.
Also, some of them propose to explore low resolution sensors~\cite{ryoo2018extreme,chou2018privacy}.
Others propose to explore defocus blur as an aberration related to the parameters of a well-designed sensor to preserve privacy~\cite{pittaluga2016pre}.
However, this solution is model-based and does not provide any insight on the use of DNNs, as proposed in this work.
More recently, in~\cite{hinojosa2021learning}, Hijonosa~\textit{et al.} makes use of Deep Optics to design an optical encoder-decoder that produces a highly defocused image that still succeeds to perform pose estimation thanks to the jointly optimized optics.
Yet, this solution is task-specific as the decoder is finetuned to fit the lens and Human Pose Estimation metrics.

In this work, we propose to study the impact of a synthetically defocused dataset on different tasks to ensure privacy and still perform well on general tasks.
To this aim, as illustrated in~\ref{fig:graphical-abstract}, we adopt semantic segmentation for a non-privacy task and License Plate Detection and Recognition as a privacy task.
In the following, we present background information on these domains.

\paragraph{Semantic segmentation} is a computer vision task in which each pixel correspond to a specific classification label. 
This challenge represents a highly competitive benchmark in CV with different approaches from DNNs which seek to better explore local and global features in a scene to improve predictions.
Long~\textit{et al.} proposed the Fully Convolutional Network (FCN) that introduced the idea of a dense pixel-wise prediction by adapting current object detection networks with an upsampling layer and a pixel-wise loss without any fully-connected layer.
This work features important improvements in the field and was followed by many contributions on the upsampling strategy to generate a high resolution output in~\cite{ronneberger2015u,badrinarayanan2017segnet,lin2017feature,Zhu_2017_ICCV,yuan2021volo}.
U-net~\cite{ronneberger2015u} is a network with an encoder-decoder architecture that introduced skip connections between the encoder and the decoder parts to improve information flow during both training and inference, while they also reduce information loss through the layers.
Chen~\textit{et al.} proposed with DeepLab series~\cite{chen2017deeplab,chen2017deeplabv3} the atrous convolution for upsampling, also known as dilated convolutions. These layers allow to enlarge the filters' view field and thus improve the objects' context information.
In DeepLabv1~\cite{chen2017deeplab}, a fully-connected Conditional Random Field (CRF) is used to capture fine details in the image. This post-processing step is eliminated in Deeplabv3~\cite{chen2017deeplabv3} while the authors also include an improved Atrous Spatial Pyramid Pooling (ASPP) module.
Latest networks such as Volo~\cite{yuan2021volo} make use of vision transformers~\cite{dosovitskiy2020image} by introducing a specialized outlook attention module to encode both context and fine-level features.

\paragraph{License Plate Detection and Recognition (LPDR)} techniques~\cite{silva2018license,gonccalves2016license,laroca2021efficient,wang2021rethinking,xu2018licenseplatedetection,alvar2022license} are widely used in applications that involve vehicle identification such as traffic surveillance, parking management, or toll control.
It typically implicates different sub-tasks, including object detection (vehicle and license plate detection), semantic segmentation (character segmentation) and finally Optical Character Recognition (OCR).
Gonçalves~\textit{et al.}~\cite{gonccalves2016license} propose to use only two steps to detect and recognize LPs and thus, reduce error propagation between sub-tasks. However, it only considers  plates with small rotations in the image plane, which is not a consistent configuration to real-world uncontrolled applications.
Notably, Silva and Jung~\cite{silva2018license} proposed a module for unconstrained scenarios divided into three phases. First, it detects vehicles using YOLOv2~\cite{redmon2016yolo9000}. Then, the cropped regions are fed into the proposed WPOD-Net, which detects license plates (LP). Next, a linear transformation is used to unwarp the LPs. Finally, an OCR network is used on the rectified LP to translate the characters into text.Later, in~\cite{silva2021flexible}, Silva and Jung extended their work by re-designing the LP detector network with specialized layers for both classification and localization.
Laroca~\textit{et al.}~\cite{laroca2021efficient} also adopts YOLOv2 as a base for an efficient model for vehicle detection. The detected regions from this network are sent to a second stage where LPs are simultaneously detected and classified with respect to their layout (\textit{e.g.}, country). Finally, they apply LP recognition based on the extra information given by the precedent step. This approach successfully works for a wide variety of LP layouts and classes.
Wang~\textit{et al.}~\cite{wang2021rethinking} propose VSNet which includes two CNNs: VertexNet for license plate detection and SCR-Net for license plate recognition. This model is capable of rectifying the position of the LP and also deals with constraints in illumination and weather.
Finally, related to our \cite{alvar2022license} proposes a annotated license plate detection and recognition version of the Cityscapes dataset. The authors also propose to apply image blurring with various levels of spread $\sigma$ to anonymize LPs. However, they propose their method as a software-level processing and do not discuss any hardware-level solution as ours.

\section{Experimental study}\label{sec:experiments}

In this section, we explain the steps followed to generate our datasets based on the Cityscapes one~\cite{Cordts2016Cityscapes}.
The goal is, while synthetically mimicking the output of a real-world privacy-designed camera, to calibrate the camera in a manner to hinders LPDR while being robust to non privacy-sensitive tasks such as the semantic segmentation of urban areas.

To this aim, we propose two main experiments: 
\begin{enumerate}[label=(\roman*)]
    \item The first explores the effect of both defocus-blurred and ``grayscaled'' input images during training and inference, where we prove the potential of a deep network to extract useful contextual information despite the imposed visual aberrations (cf. Section~\ref{sec:semseg});
    \item The second compares the generalization capability of two deep neural models, trained with standard colored and all-in-focus images, on a privacy sensitive task and a non-privacy sensitive one (Section~\ref{sec:lpdr}).
\end{enumerate}

\subsection{\label{sec:implementation} Experimental setup}

\paragraph{Dataset}
We make use of Cityscapes, a large-scale dataset for road scene understanding, composed of sequences of stereo images recorded in 50 different cities located in Germany and neighboring countries.
The dataset contains 5000 densely annotated frames with semantic, instance segmentation, and disparity maps.
It is split into 2975 images for training, 500 for validation, and 1525 for testing.
Test results are computed based on its validation subset, as we do not have access to the testing ground-truth.

\begin{figure}[htb]
  \centering
  \includegraphics[width=1.0\textwidth]{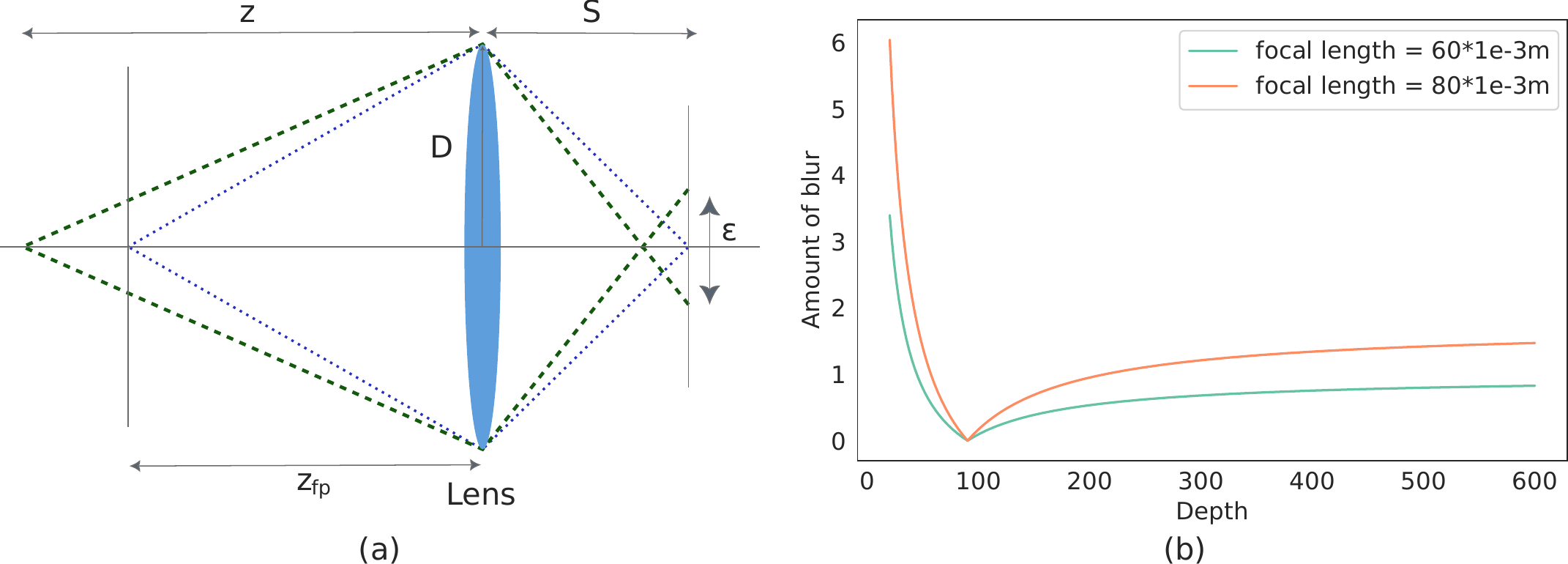}
  \caption{\label{fig:defocus-blur-lens} 
  Illustration of the principle of defocus blur. (a) A point source image placed in a distance, $z$, out of the in-focus plan position, $z_{fp}$, results on an ammount of blur, $\epsilon$, which is variable with respect to $z$. (b) Defocus blur variation with respect to depth and two focal length values.
  }
\end{figure}

\paragraph{Defocus Blur}
In contrast of post-processing techniques where identifiable features are blurred when detected, here we simulate a realistic hardware-level defocus blur to ensure PbD.
Indeed, defocus blur is an optical aberration inherent to a camera and its parameters, illustrated in Figure~\ref{fig:defocus-blur-lens}.

In Section~\ref{sec:semseg}, we chose parameters that correspond to a synthetic camera with a focal length of $80~mm$, a pixel size of $4.4~\mu m$ and an f-number of $2.8$.
The later refers to the ratio of a lens' focal length to its aperture's diameter. It indicates the ammount of light coming through the lens. Thus, we set it to a very low value, so the simulated camera has a very shallow depth of focus and is very sensitive to defocus.
The choice of a long focal length corresponds to the desired shallow depth of field.
In comparison, in Section~\ref{sec:lpdr}, we adopt a focal length of $60~mm$ to soften the amount of defocus blur through the image.

In Figure~\ref{fig:defocus-blur-lens}.b, we show the variation of blur with respect to depth from the sensor. We show two functions corresponding to both aforementioned configurations. As stated, at all depths but in the in-focus position, keeping all other parameters intact, a larger focal length correspond to more intense blur.

To generate synthetically realistic defocus blur, we apply the approximate layered approach proposed by Hasinoff~\textit{et al.} in~\cite{hasinoff2007layer} and use the implementation proposed in~\cite{carvalho2018defocus}.
This method consists on the sum of $K$ layers of blurred images, corresponding to different depths in the image and the respective blur ammount for each $z$.
In such manner, we are able to generate depth maps from the provided disparity maps, as well as camera intrinsic parameters by triangulation: 
$z = {(f*b)}/{d},$
where $z$ is the depth map, $f$ the focal length, $b$ the baseline, and $d$, the disparity map.

The Cityscapes provided disparity maps~\cite{Cordts2016Cityscapes} have missing values that can be related to a fault in the stereo correspondence algorithm~\cite{merrouche2020objective}, or occlusions.
Hence, when generating the depth maps, the corresponding pixels are set to zero.
By extension, these pixels correspond to increased blur in some areas of the image, as closer objects to the camera present higher blur kernel diameters.
This makes an accurate semantic segmentation more challenging.

\paragraph{Semantic Segmentation}
We use as a benchmark to perform multi-class semantic segmentation, DeepLabv3 with a ResNet-101 as a backbone for feature extraction.
To avoid cross-knowledge from colored or in-focus images, we do not initialize our model with pretrained weights on other datasets, as it is commonly done in the field.

\subsection{\label{sec:semseg}Semantic segmentation sensitivity to color and defocus blur}

In this set of experiments, we explore the effects of visual quality on the robustness of a semantic segmentation deep model.
To this aim, we carefully tackle two types of quality distortions that can be related to the hardware limitations of an existing camera:
\begin{itemize}
    \item defocus blur (B): an optical aberration related to the parameters of the camera and the position of the focal plane;
    \item ``grayscaling'' (G): a transformation that mimics a monochromatic camera.
\end{itemize}

This results in three new different versions of Cityscapes:
\begin{itemize}
    \item Cityscapes-B: contains physically realistic defocus blur;
    \item Cityscapes-G: contains only grayscale images;
    \item Cityscapes-BG: contains both transformations.
\end{itemize}

We train the DeepLabv3 model for semantic segmentation on all four (3 variations and the original) versions of the dataset.
Finally, we compare the resulting trained models across different input information with and without quality deformations.

\begin{figure}[htb]
  \centering
  \includegraphics[width=1.0\textwidth]{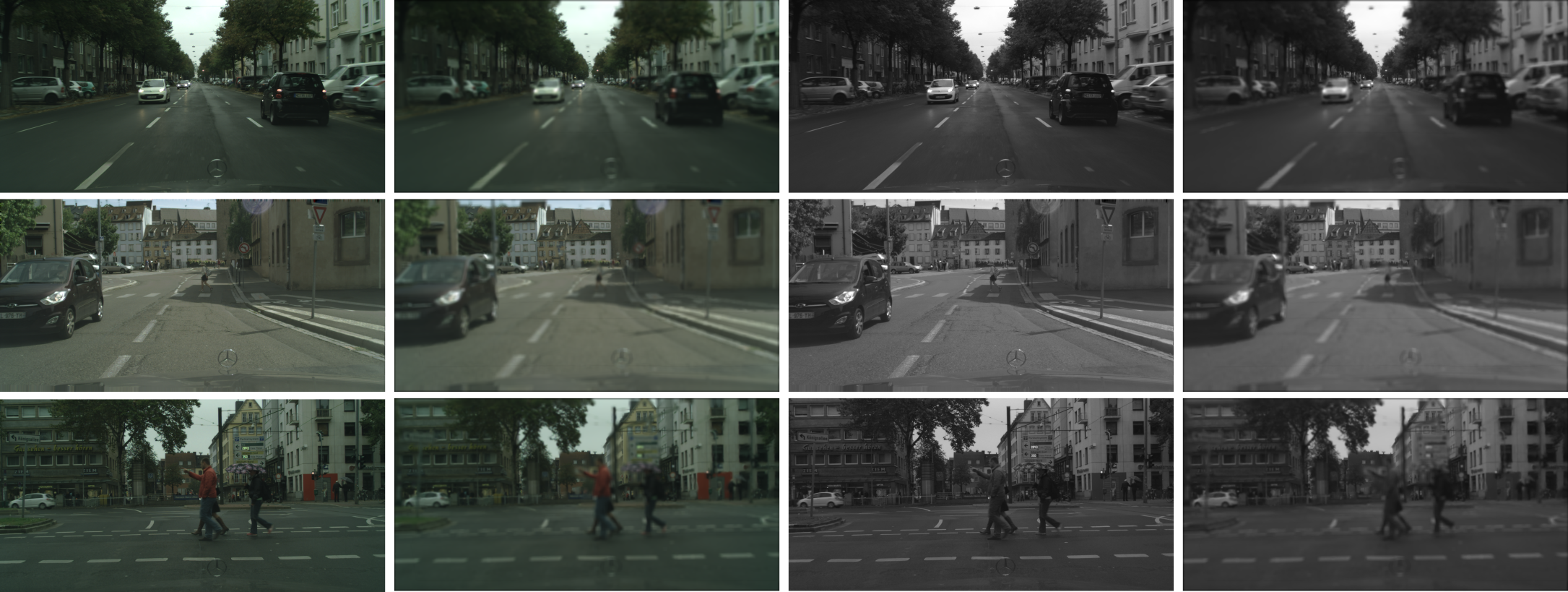}
  \caption{\label{fig:corrupted-images} Samples from Cityscapes with the applied distortions. 
  First (\textit{resp.} third) column has all-in-focus RGB (\textit{resp.} grayscale) images.
  Second (\textit{resp.} forth) column has out-of-focus RGB (\textit{resp.} grayscale) images with a focal plane at 400m.
  }
\end{figure}

Cityscapes-B is generated according to~\cite{hasinoff2007layer,carvalho2018defocus}, as explained in Section~\ref{sec:implementation}.
Our main goal is to reduce the sensitive information that is often related to high-frequency such as facial features, small personal objects, or identifiable texts (\textit{e.g.}, license plates).
To achieve this, the defocus blur is made more intense for pixels that account for objects near the camera. 
Note that the ones which are far away from the sensor have lower resolution information which already compromises enough their sensitive data. By exploring the defocus blur, we have more attenuated distortion in far points. This allows to extract more information from objects in these positions, with respect to models that blur the whole image with the same kernel.
Accordingly, we set the focal plane to a distant point, more precisely, to $400 m$.
Therefore, we generate out-of-focus images like the ones illustrated in the second row of Figure~\ref{fig:corrupted-images}.

Additionally, to generate Cityscapes-G, we convert color images to grayscale and we copy the single-channel information to generate an input with the same number of channels as a RGB image. This step was performed to maintain the same structure of the deep model.

Other image transformations include resizing all input images from $1024\times2048$ to $512\times512$ during training and testing; and applying random horizontal flip and small rotations for data augmentation. Therefore, we upsample the generated semantic segmentation maps using the nearest neighbor resampling method to the original size while producing metrics.

\begin{figure}[h!]
  \centering
  \ifcompilegraphics
  \includegraphics[width=.78\textwidth]{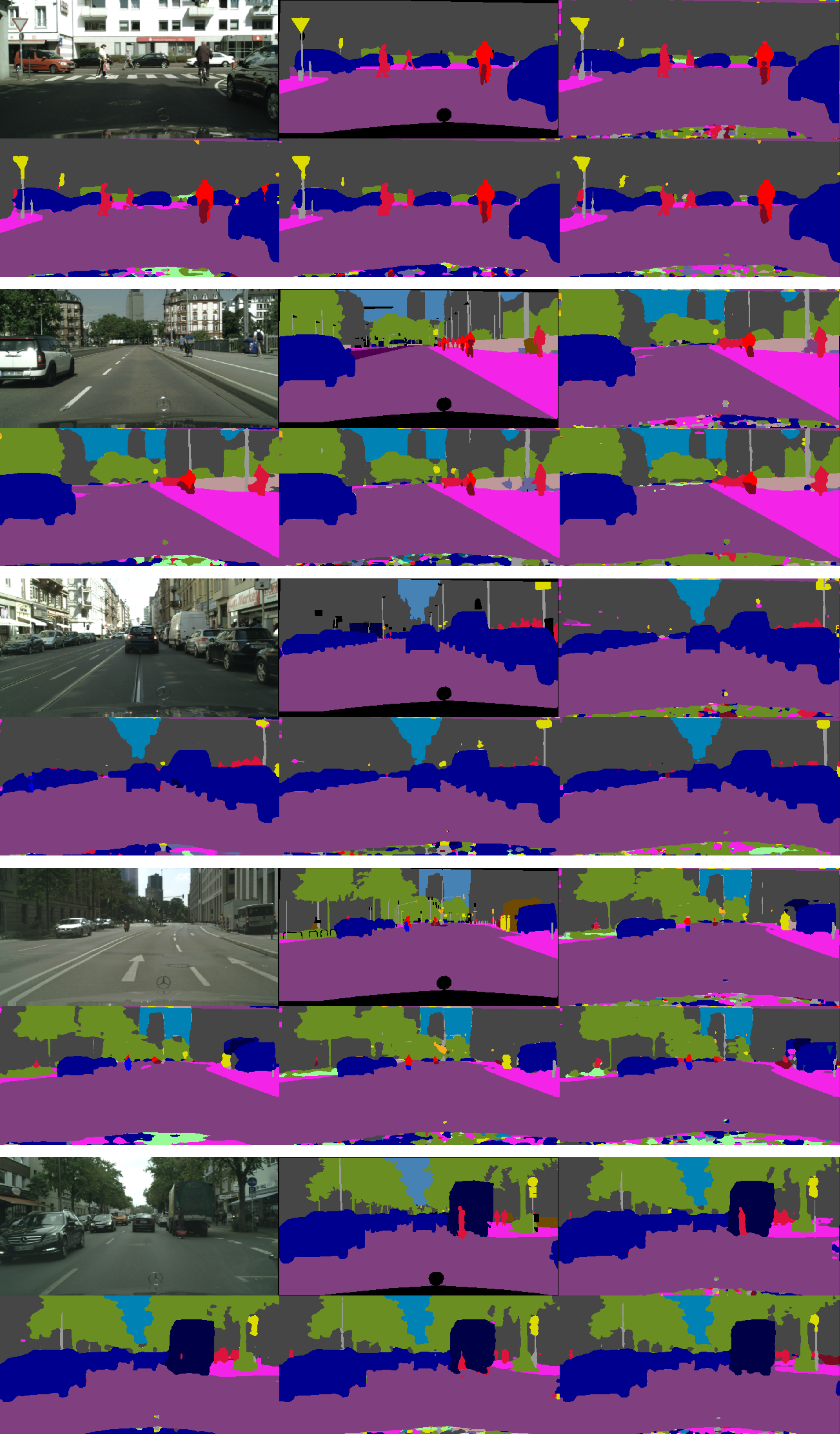}
  \else
  \includegraphics[width=.6\textwidth]{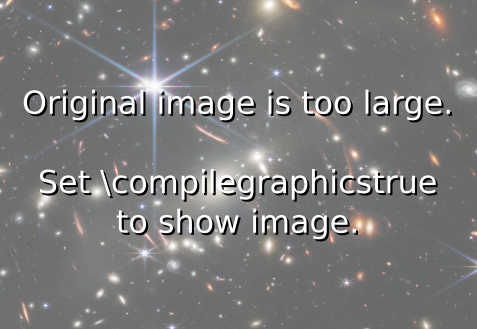}
  \fi
  \caption{\label{fig:semseg400_visual} Qualitative results of semantic segmentation on distorted versions of Cityscapes. For each sample, we have, original RGB, ground truth; and models' outputs when trained and tested on: Cityscapes-C, Cityscapes-B, Cityscapes-G and Cityscapes-BG. These models correspond to the ones with highlighted scores in Table~\ref{table:semseg}.
  }
\end{figure}

Finally, we trained our model with each one of the four versions of Cityscapes mentioned in the beginning of this section and performed inference metrics also on all versions for each model.
This allows us to analyse the out-of-distribution robustness of the network faced to different visual qualities.
The scores of our models are shown in Table~\ref{table:semseg} and we have some qualitative examples in Figure~\ref{fig:semseg400_visual}.
They are grouped into four ensemble of rows which correspond to each trained model.
The second and third column indicates if the model was trained with grayscale (G), defocus blur (B) deformations, both or none.
Following a similar pattern, the first column indicates the dataset related to the metrics during inference.
For instance, C means that the dataset is the original colored one (C).

\begin{table}[htb]
  \centering
  \small
  \caption{
    \label{table:semseg} Comparison of Cityscapes-\{C,G,CB,GB\} trained on DeepLabv3 for semantic segmentation.
    We highlight in grey the results where models were trained and tested with the same dataset version.
  }
  \begin{tabular}{cccccccc}
    \hline \noalign{\smallskip}
                  Test
                   & \multicolumn{2}{c}{Train} & \multicolumn{3}{c}{Metrics}\\
                  & G & B & mA$\uparrow$ & mIOU$\uparrow$ & OA$\uparrow$\\

    \hline \noalign{\smallskip}
\rowcolor{lightgray}
    C   &  &     & 0.738 & 0.649 & 0.939    \\
    CB  & &      & 0.592 & 0.489 & 0.900    \\
    G   & &      & 0.628 & 0.526 & 0.908    \\
    GB & &       & 0.388 & 0.282 & 0.799    \\
    
    \hline \noalign{\smallskip}
    C   & & \cmark    & 0.502 & 0.401 & 0.879      \\
    \rowcolor{lightgray}
    CB  & & \cmark    & 0.716 & 0.611 & 0.929      \\
    G   & & \cmark    & 0.381 & 0.275 & 0.802      \\
    GB  & & \cmark    & 0.568 & 0.436 & 0.883      \\

    \hline \noalign{\smallskip}
    C   & \cmark &    & 0.708 & 0.615 & 0.930      \\
    CB  & \cmark &    & 0.455 & 0.371 & 0.855      \\
    \rowcolor{lightgray}
    G   & \cmark &    & 0.730 & 0.640 & 0.933      \\
    GB  & \cmark &    & 0.463 & 0.379 & 0.853      \\

    \hline \noalign{\smallskip}
    C   & \cmark & \cmark    & 0.380 & 0.319 & 0.860      \\
    CB  & \cmark & \cmark    & 0.644 & 0.550 & 0.915      \\
    G   & \cmark & \cmark    & 0.412 & 0.347 & 0.870      \\
    \rowcolor{lightgray}
    GB  & \cmark & \cmark    & 0.664 & 0.569 & 0.917      \\

    \noalign{\smallskip}
    \hline
  \end{tabular}
\end{table}

From the results, we first compare the models trained and tested on the same version of the dataset and next we discuss about their generalization capability to out-of-distribution data. The highlighted models in Table~\ref{table:semseg} show that using both distortions hurt more the performance than using each one of them alone. However, there is no significant drop in metrics.      Also, the outputs illustrated in Figure~\ref{fig:semseg400_visual} present very simular results from different inputs. Still, we chose some images that present notable effects of defocus blur and grayscale transformation. In general, the presence of blur dammage the perception of object's boundaries. This explains loss in fine details and missing detections of details. While missing colors may help fusion some parts of the scene as we can notice in the fifth row, where the man near the vehicle disappears for the GB model. This situation can explain some of the loss in accuracy, but can be explained away with richer data augmentation or even exploring better the high resolution input images.

At this point, we design a compromise between gain in privacy against a small loss in accuracy for this type of task. The present experiment shows that at the same time these visual distortions influence on the model's capability to extract important information, they do not significantly weaken the model's performance.

In addition, we can also observe how each model generalize information when faced to out-of-distribution data. The biggest drop in accuracy happens to the model trained with Cityscapes-C, when using BG images as input during test. Indeed, the applied transformations reduce significantly the efficiency of the network as already discussed in~\cite{dodge2016understanding,zhou2017classification,hendrycks2018benchmarking}. This means the network also needs to seem these distorsed types of image to improve generalization. \\

In this section, we showed how defocus blur and grayscale can be adopted as visual aberrations to improve privacy and still allow a semantic segmentation model to perform well. In the next section, we will also observe the behavior of a privacy-sensitive task, LPDR.

\subsection{\label{sec:lpdr}LPDR sensitivity to defocus blur}

\begin{figure}
  \centering
  \includegraphics[width=.7\textwidth]{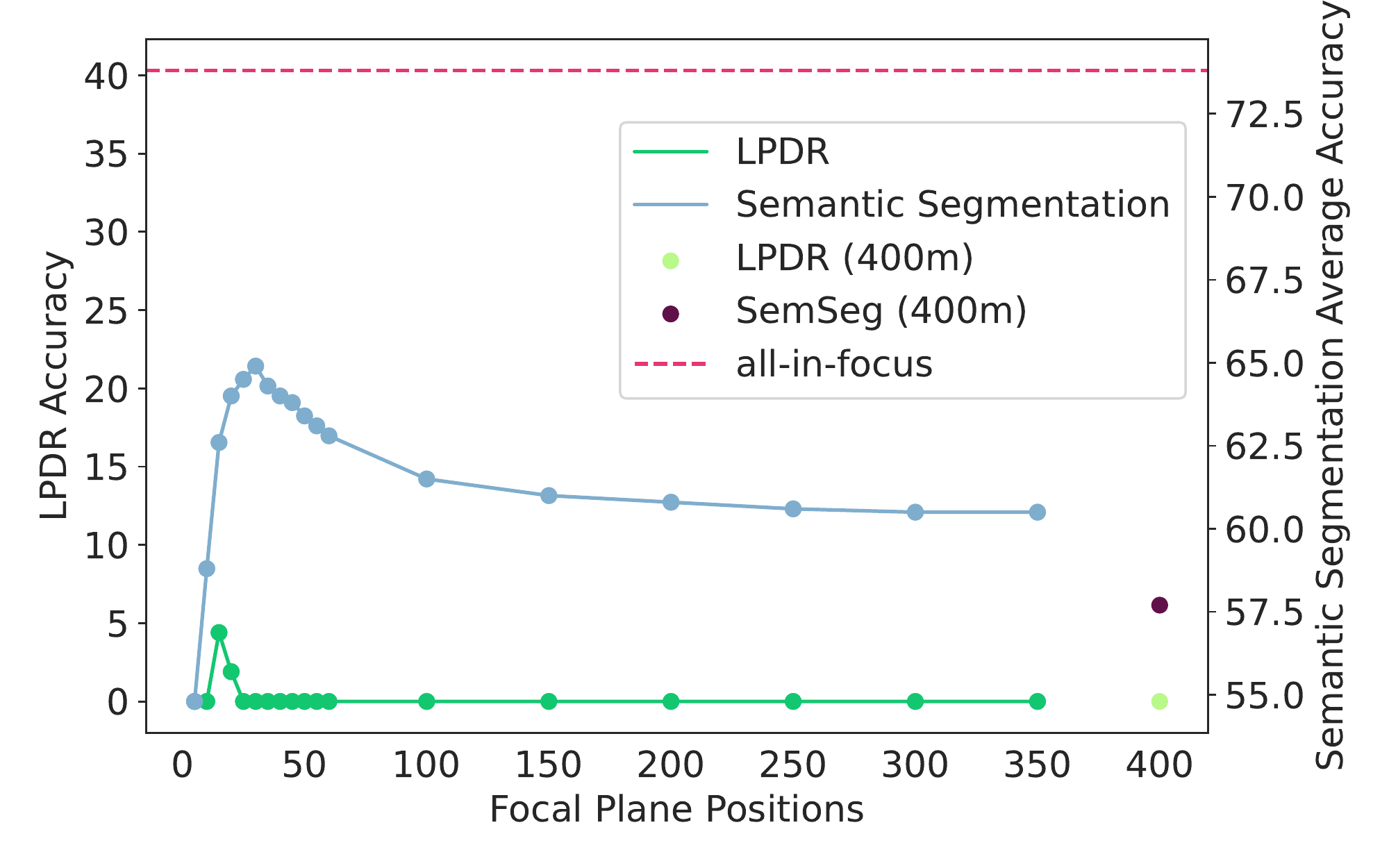}
  \caption{\label{fig:2nd-exp-blurvsperformance} 
  Quantitative results on the robustness of the models with respect to different levels of blur amount.} 
\end{figure}

We now conduct our analysis on the performance of both a semantic segmentation model and an LPDR model to different levels of severity of defocus blur with respect to the distance of the camera. To this aim, we apply the same method described in Section~\ref{sec:semseg} to generate out-of-focus images. However, here we also vary the positions of the in-focus plane to the positions illustrated in Figure~\ref{fig:2nd-exp-blurvsperformance}, and we reduce the focal length value to lessen corruption with respect to the latest experiment.

\paragraph{Models} From the previous section, we adopt the model trained with the original Cityscapes and for the task of LPDR, we use the model proposed in~\cite{silva2018license} with the original parameters.

\paragraph{Cityscapes-LP} Alvar~\textit{et al.}\cite{alvar2022license} proposed license plate annotations for 121 images from the validation subset of Cityscapes. These images were chosen with respect to the previous detection of at least one readable LP in each sample.

To compare our models, we chose only one metric for each task. For semantic segmentation, we pick the average accuracy and for LP recognition, we calculate the accuracy only considering the percentage of correctly recognized LPs. We do not generate metrics for the object detection step in LPDR.

\begin{figure}
  \centering
  \ifcompilegraphics 
  \includegraphics[width=.9\textwidth]{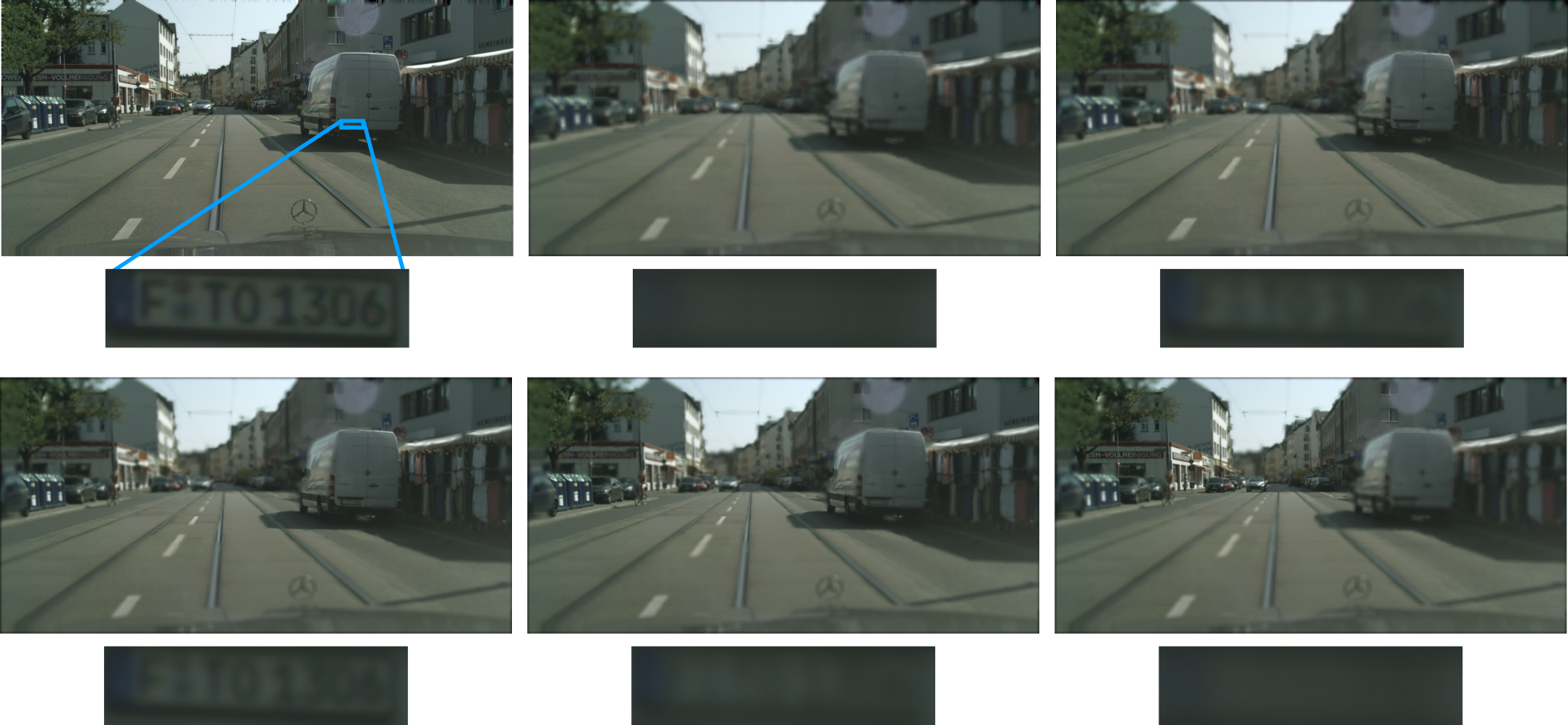}
  \else
  \includegraphics[width=.6\textwidth]{generic_large_image_info.png}
  \fi
  \caption{\label{fig:2nd-exp-defocus-blur-different-focal-planes-on-images} Effect of different blur severities with respect to the defocus blur and the variation of the focal plane position.}
\end{figure}

These results are illustrated in Figure~\ref{fig:2nd-exp-defocus-blur-different-focal-planes-on-images}. We add the accuracies for all-in-focus images as a dashed line for both tasks and also those for a out-of-focus Cityscapes-LP with camera and in-focus plane parameters from previous experiment.

These images represent a great challenge to the LPDR model as the LPs appear in different positions, rotations and qualities. The best model manages to find approximatelly 40\% of the annotated LPs, however, the best model on out-of-focus data can only succeed at near 4\% when in-focus plane is at 15 meters. Let that be clear that a privacy-aware model should have an in-focus plan in a position where identification of personal information should be impossible because of image resolution, for example. So, even when we have small amounts of blur in a near position to the camera, the model fails.

On the other side, the segmentation model achieves much better scores even though the network was trained with only all-in-focus images. What explains that this results are better from those in the last section is the choice of a smaller focal length, as already mentioned and illustrated in Figure~\ref{fig:defocus-blur-lens}b. This renders the model less sensitive to defocus blur. We made this choice for sake of fairness to compare the influence of defocus blur on LPDR.

\section{Conclusions}

In this paper, we raise some privacy concerns related to the increasing exploitation of images in the public space.
Hence, we discuss the principles of Privacy by Design, which stands that a product should respect consumers personal information since its very own conception.
Finally, we propose an approach to perform hardware-level anonymization by using defocus blur and grayscale images as inputs to neural networks.
We show through our experiments that it is possible to improve privacy without harming more general non sensitive tasks like semantic segmentation.
The advantage of our module is that it presents a general approach which can also be extended to other CV tasks such as object detection or classification.
To the best of our knowledge, our work is the first to propose coupling defocus blur and a DNN model to strategically hide identifiable information without compromising concurrent tasks of more general purposes. 
Future directions for experimenting on more efficient models of different tasks, increasing data augmentation and conceptualizing the co-design of a privacy-aware camera to perform tasks with DNNs.

\bibliographystyle{splncs04}
\bibliography{egbib}
\end{document}